# CLUSTERING AN AFRICAN HAIRSTYLE DATASET USING PCA AND K-MEANS

Teffo Phomolo Nicrocia, Owolawi Pius Adewale, Pholo Moanda Diana

Department of Computer Engineering, Tshwane University of Technology, Pretoria (Soshanguve)

## ABSTRACT

*The adoption of digital transformation was not expressed in building an African face shape classifier. In this paper, an approach is presented that uses k-means to classify African women images. African women rely on beauty standards recommendations, personal preference, or the newest trends in hairstyles to decide on the appropriate hairstyle for them. In this paper, an approach is presented that uses K-means clustering to classify African women's images. In order to identify potential facial clusters, Haarcascade is used for feature-based training, and K-means clustering is applied for image classification.*

## KEYWORDS

*Face detection, k-means, African, hairstyle*

## 1. INTRODUCTION

Hairstyles have always been an expression of beauty and personality for African women. In many cultures, people see a woman's hair as her crown of glory and consider it a lordly symbol that symbolizes her femininity. In the early 15th century, for example, hair served as a carrier of messages in West African countries such as Yoruba (Nigeria) and Mandingo (Sierra Leone) communities; it communicated marital status, ethnic identity, wealth, and status in the community while also identifying geographical origin [1]. Historical background and the influence of mainstream media, that promotes beauty stereotypes that esteem European standards, have affected how African women perceive their natural hair [2]. As highlighted by Bankhead & Johnson [1], it is therefore very important to educate young girls and women about accepting their natural hair and to motivate African women to view positive images of themselves. According to psychologists, having beautiful hair and the feeling of having such an appearance directly affects the psychology of women in a positive or negative way .In addition , a good hairstyle can highlight facial features and can help improve facial appearance by 70% as it has the tendency to enhance facial features [10].However, as we alter our hairstyle to embrace our authentic selves, we have to remember that every person is different. It is necessary to match the characteristics of each individual with a suitable hairstyle [3]. Many hairstyle recommendation systems exist, as shown in Table 1. However, most of them have been created using datasets that do not accommodate the of African women. In the existing datasets, mainly Caucasian faces are present, while our main focus is on the classification of African women's faces.

Table 1. Related Work

| Paper | Face Shape classification method | Target |
|---|---|---|
| Alzahrani, Al-Nuaimy & Al-Bander (2021) | Inception V3 | Men and Women |
| Chen,Zhang,Huang, Chen&Luo (2021) | YOLO v4 | Men and Women |
| Weerasinghe & Vidanagama (2020) | Haarcascade Classifier | Men and women |
| Liu, Ji, Hu & Zhan (2019) | VGGNet-16 | Men and women |
| Pasupa, Sunhem & Loo (2019) | Support Vector Machines | Women |
| Rajapaksha & Kumara (2018) | Dlib machine learning library (C++) | Men and Women |

Additionally, most of these studies use the five classic face shapes (oval, heart, round, oblong and square) to classify facial features. In this paper, we will explore the gathered images using clustering methods to see if these five face classes are a fit for our dataset. We will then classify each of the faces into one of the clusters obtained, as the basis for an African hairstyle recommendation system.

This study aims to accomplish the following major objectives:

- Collect images of African women to construct a dataset;
- Extract facial features of collected images using Haar Cascade method;
- Using K-Means clustering to perform image classification to identify the potential facial clusters(groupings);
- Develop and evaluate a multi-layer perceptron for assigning each image to a cluster.

## 2. MATERIAL AND METHODS

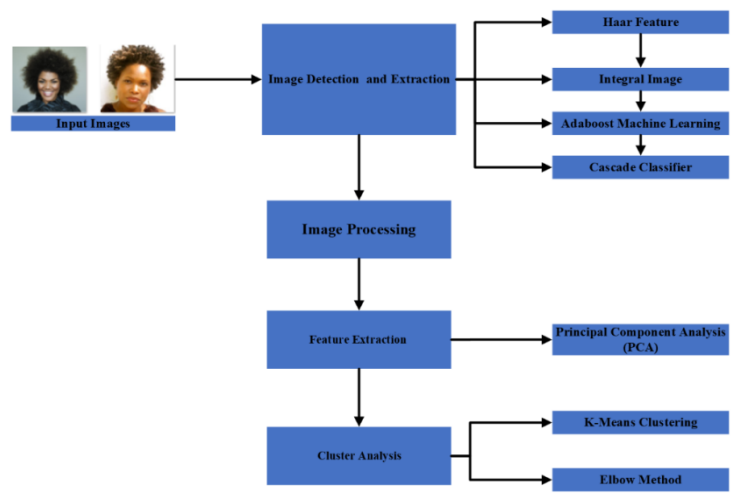

Figure 1. Block diagram of clustering images

## 2.1. Data Collection

The image datasets are obtained from Figaro1K dataset and include images gathered from the web for scientific research conducted by non-profits. There are six classes of hair in the Figaro1K database, which are: straight, wavy, curly, kinky, braids, dreadlocks, and short men, each containing 120 images. African women typically have afro-textured hair or kinky hair, so I selected images based on that hair type.

## 2.2. Facial Detection and Extraction

Human faces generally share a similar geometrical configuration while sharing the following features: eyes, mouth, nose, and chin. The Viola-Jones face detection algorithm goes through an image to detect these features. The method uses Haar-like features to extract the face and construct a bounding box around it. These features are filters run through an image to determine the location of the face [5].

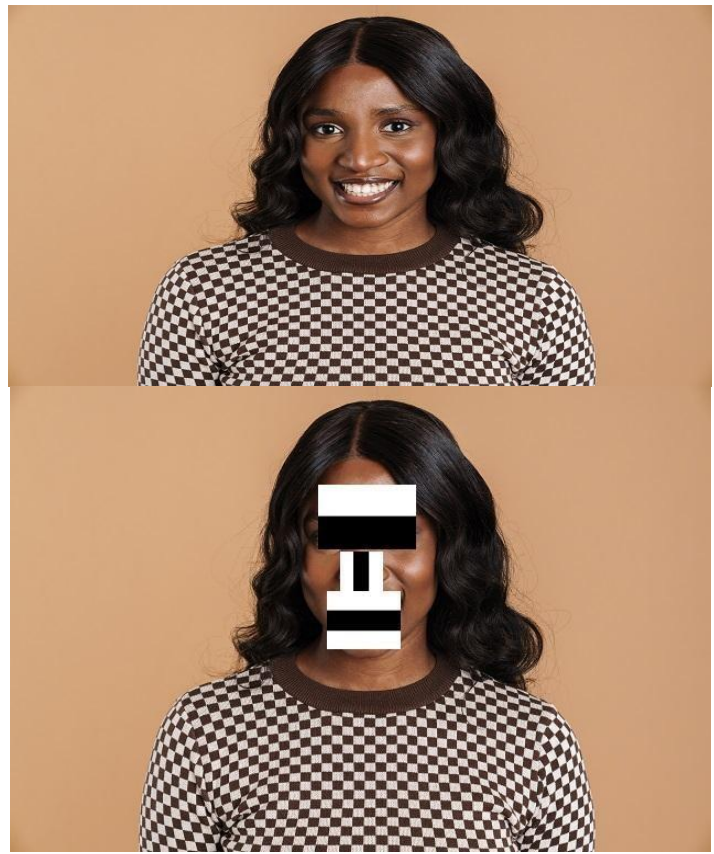

Figure 2. Block diagram of clustering images

### 2.2.1. Calculating Haar Values

When scanning the image, the Haar value of each region is calculated as follows [6]:

$$HaarValue = \frac{Sum\ of\ dark\ pixels}{Number\ of\ dark\ pixels} - \frac{Sum\ of\ light\ pixels\ values}{Number\ of\ light\ pixels}$$

To make the calculations faster, images can be converted to integral images. To obtain an integral image, the value of each pixel is replaced by the sum of all pixel values above it, all pixel values to its left and the value of the pixel itself [5]. Using the integral image, the haar value is computed using only four values for each region, as shown on Figure below.

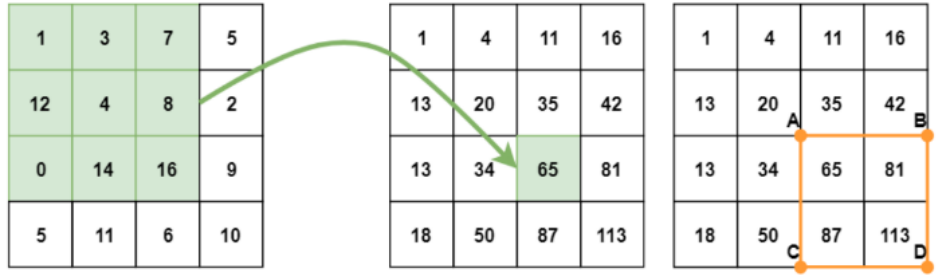

Figure 3. Integral image

## 2.2.2. AdaBoost Learning Algorithm

The majority of the Haar features obtained after are irrelevant and random. To select a good subset of features from a huge set of features, the AdaBoost algorithm is applied [16]. AdaBoost classifies whether a weighted combination of the Haar features contains a face or not, and so reduces the number of potential facial regions [5][7]. The classifiers have excellent feature selection mechanisms and fast learning rates, which makes them superior to other machine learning algorithms.

## 2.2.3. Cascade Filter

The AdaBoost algorithm returns 6,000 features which are still non-face regions in majority [19]. As a next step, the algorithm therefore implements cascading classifiers. These latter are a series of decision trees applied to the image regions obtained from the previous step. Only image regions that are classified as facial regions by all cascading classifiers are ultimately retained [7]. This minimizes the algorithm's false negative rate.

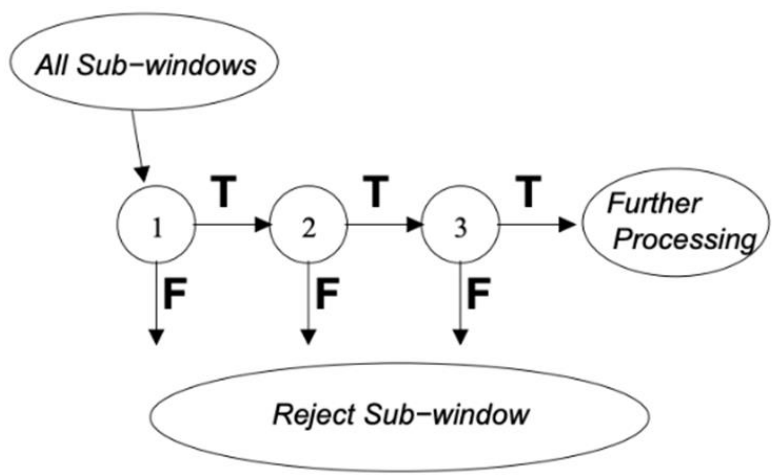

Figure 4. Cascade Classifier

## 2.3. Image Processing

Image processing is the first step when supplying image data to any algorithm. The following steps were involved in this stage image processing:

- Resizing: All extracted faces were reduced to the same size: 224 by 224 pixels.
- Colorscale: The images were then converted to grayscale (2-D).
- Scaling: All pixels should be normalized between 0 and 255.
- Flattening: in preparation for the next steps, the images were converted from 2-dimensional arrays of 224 by 224 pixels into 1-dimensional arrays of 50176 pixel.

## 2.4. Feature Extraction

Principal component analysis involves extracting features from an image, visualizing complex data, and capturing variation in the data. Principal Component Analysis aims to reduce the dimensions of a dataset by selecting uncorrelated features that capture the most information about the data [8]. The algorithm for Principal Component Analysis is as follows [11].

Step 1:Reduce continuous features to the same scale

$$zi = \frac{(xi - \mu)}{\sigma}$$

Step 2: Generate the covariance matrix

Covariance expresses the relationship between two variables and is calculated as ffollows:

$$COV(X, Y) = \frac{\sum (Xi - X)(Xi - Y')}{n}$$

where x' = mean of variable X; y' = mean of variable Y; n = number of data points. The covariance is calculated between every pair of variables and the results are saved in a p x p matrix (p = number of variables in the dataset).

Step 3: Calculate eigenvectors and eigenvalues

For a square matrix A, the eigenvector (v) and eigenvalue (λ) are obtained by solving the following equation:
$Av = \lambda v$
Step 4: Create a feature vector to decide which principal components to keep.

Step 5: Recast the data along the principal components' axes.

## 2.5. Cluster Analysis

### 2.5.1. K-Means Clustering

The K-Means algorithm is the most commonly used partitioning-based clustering algorithm. Below is a brief outline of the K-Means algorithm [17]:

- Randomly select k points and make them the initial centroids.

- Choose an element from the collection, compare it with each centroid, and then assign that element to the cluster that corresponds to that centroid.
- Recalculate the centroids after each data point is assigned to a cluster.
- Steps 2 and 3 should be repeated until no data point moves from its previous cluster to some other cluster.

A variety of methods can be applied to estimate the optimal number of clusters k based on various parameters. This study measures the quality of clusters by using the Elbow Method and the silhouette score to determine the optimal number of clusters k.

### 2.5.1.1. Elbow Method

The elbow method involves comparing the number of clusters that will form an elbow on the curve to determine the optimal number of clusters. A cluster value is considered best if its value forms an elbow on the curve. To compare cluster values, calculate SSE (Sum of Square Error). Consequently, if there are more clusters K, then the SSE (Sum of Squared errors) value will be smaller since the number of clusters is greater [18]. The downside of the method is that when k is increased, the SSE value will approach zero, which makes it impossible to determine if a cluster is good or bad [13]. To overcome this drawback, this method suggests that k should be calculated when the maximum error value of the graph reduces significantly [14].

### 2.5.1.2. Silhouette Score

The silhouette score was developed as a means of exploring and analyzing the distance between the resulting clusters. In addition to that, it also serves as a measure of how well a cluster is doing over a given period. In Silhouette, a score varies from -1 to 1 and -1 indicates that the time-series was assigned to the incorrect cluster, 0 indicates that there are clusters that overlap with samples near neighboring clusters and 1 indicates that it belonged to the correct cluster [18]. For each sample, to calculate the Silhouette score, we use the following formula:

$$S(X) = \frac{(BX - AX)}{\max(AX, BX)}$$

- Assuming that X object is in a cluster, calculate its average distance from each other object in the cluster. Let AX be the value of this parameter.

- Prior to finding its average distance from X, find its average distance from all other clusters containing objects in the same cluster. Calculate the minimum value based on all the clusters. Let BX be the value.
- The silhouette value of the X object is:

$$S(X) = \frac{(BX - AX)}{\max(AX, BX)}$$

## 3. RESULTS AND DISCUSSION

From the Figaro1K dataset, we extracted only 112 images that contain only African women from the total of 672 images.

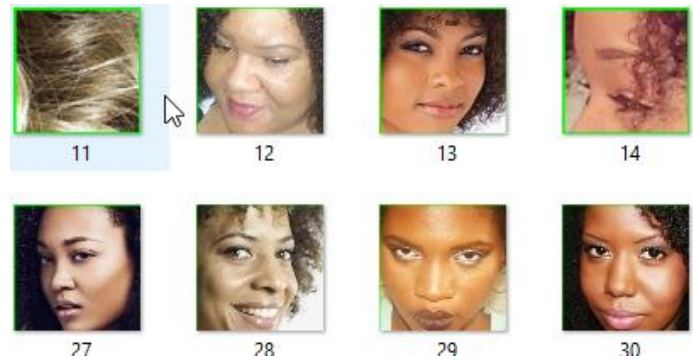

Figure 5. Object detection by Haar Cascade Classifier

As shown in Figure 5, the Haar cascade classification results are displayed. The Haar cascade classification was trained on images that were not identical regarding brightness of the faces and lack of sharp features. This resulted in the faces not being spotted and caused the classifier to struggle to classify the images. The algorithm's average recall is low due to many false positives and negatives, which indicates a low accuracy rate.

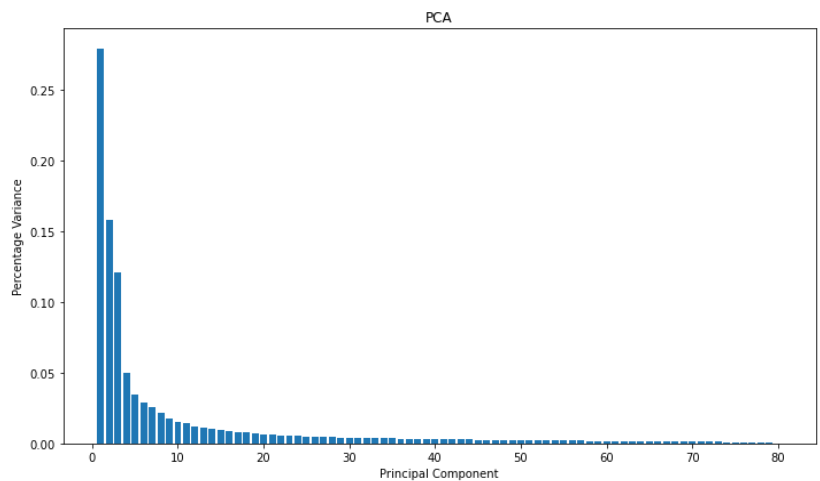

Figure 6. Principal Component Analysis

As shown in figure 6, the first 80 Principal Components explained 99.99% of data variance.

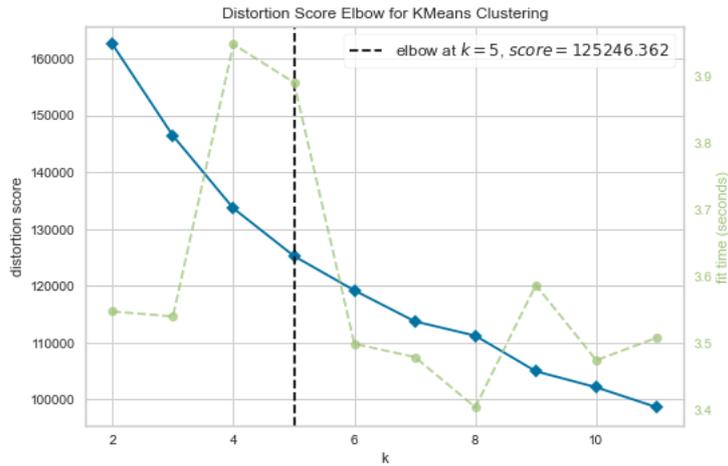

Figure 7. Elbow Method

In Figure 7, we see that there is no clear elbow. However, there seems to be a bend around k = 5. Therefore, we looked at the silhouette score for k values around 5. When looking at Figure 8, we see that k = 4 is a good choice since the Silhouette scores of all four clusters are higher than the average score of the entire dataset (indicated by the red vertical line). Additionally, the sizes of the clusters are more similar with k = 4 than with k = 3, 5 or 6.

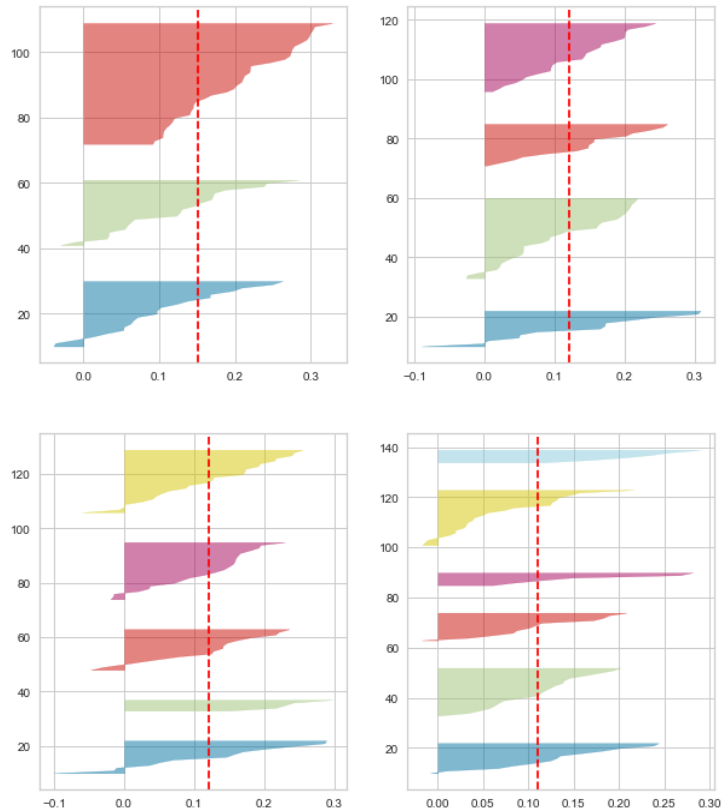

Figure 8. Silhouette Score

Below are some of the figures obtained in each cluster.

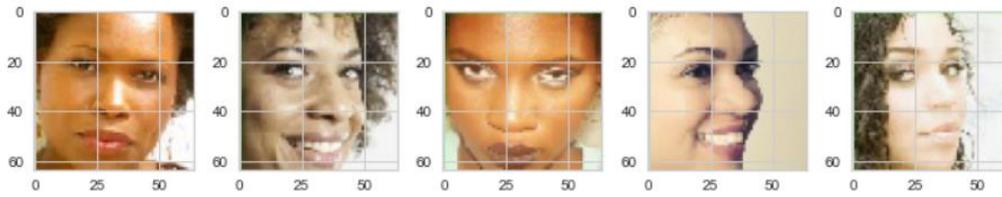

Figure 9. Cluster 0

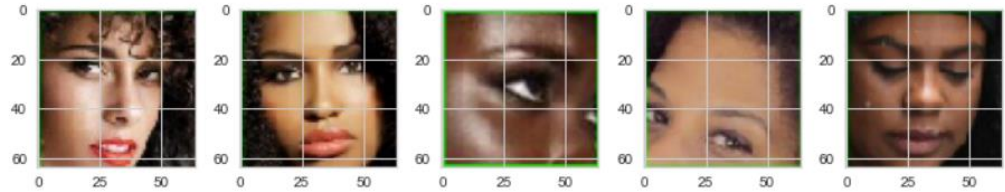

Figure 10. Cluster 1

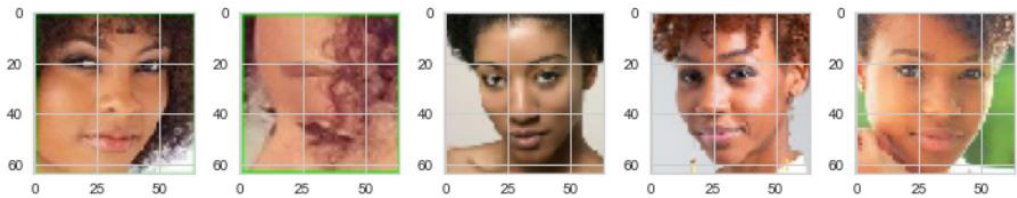

Figure 11. Cluster 2

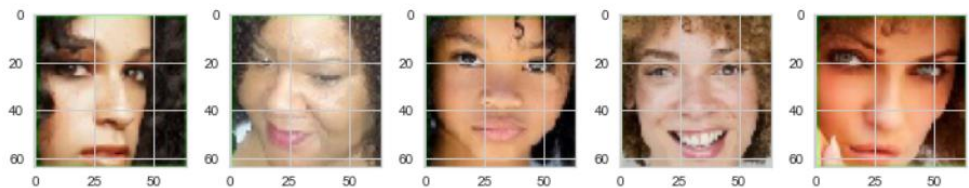

Figure 12. Cluster 3

## 4. CONCLUSION

In this paper, we have examined the face of African women from the Figaro1K dataset. After extracting facial features using Haar cascade classifiers, we performed feature selection using Principal Component Analysis. To identify the potential facial groups or clusters, we used K-Means for clustering and obtained four clusters. Using Haar cascade yielded many false negatives and false positives and reduced the number of faces used for clustering. Future research will therefore focus on extracting more faces using more advanced algorithms. of the new dataset will then be used for creating a recommendation system that first assigns an image to a cluster then recommends hairstyles based on visual similarity.

## AUTHORS

The author, **Phomolo N. Teffo**, was born in Limpopo (South Africa) in 1993 and is currently pursuing a master's degree at Tshwane University of Technology. She has a bachelor's degree in intelligent industrial systems from the same university. Currently, she is employed by the same university as a part-time lecturer.

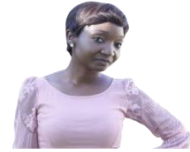

**Pius Adewale Owolawi** studied electrical engineering at the Federal University of Technology, Akure, Nigeria, in 2001, and at the University of Kwazulu Natal, South Africa, in 2006 and 2010. Tshwane University of Technology, South Africa, currently employs him as Department Head of Computer Systems Engineering. Among his research interests are RF, green communication, radiowave propagation (microwave/millimeter wave systems), satellite and free space optical communications, the Internet of Things, embedded systems, machine learning, and data analytics. In 2012, he was awarded the Joint Holder of Best Paper Award for a paper he presented at the 2nd International Conference on Applied and Theoretical Information Systems Research, which was held in Taipei, Taiwan, in 2012, and in 2015 he was awarded the Vice Chancellor's Teaching Excellence Award.

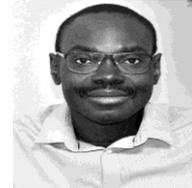

The PhD candidate **Moanda D.Pholo** was born in Kinshasa (DRC) in 1987. In 2014, she earned a Master's degree in Intelligent Industrial Systems from the same university, as well as an MSc in electronics from the Ecole Supérieure d'Ingénieurs en Electrotechnique et Electronique. Currently, she works as a data scientist in Pretoria, South Africa. She has published several scientific articles, including one in the GlobalSIP 2019 Proceedings. In Italy, Mrs Pholo recently received the "Best work in the category of AI/big data" award at the 3rd ICEHTMC.

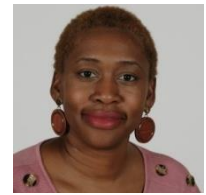